\documentclass{article}

\usepackage[preprint]{corl_2022} 

\usepackage{alltt}
\usepackage{caption}
\usepackage{paralist}
\usepackage{nicefrac}
\usepackage{booktabs}
\usepackage{algorithm}
\usepackage{algorithmic}
\usepackage{xspace}
\usepackage{amsmath}
\usepackage{svg}
\usepackage{marvosym}
\usepackage{wasysym}
\usepackage{verbatim}
\usepackage{subfigure}
\usepackage{hyperref}
\usepackage{multirow}
\usepackage{cite}
\usepackage[capitalize]{cleveref}
\usepackage[per-mode=symbol,detect-all]{siunitx}
\usepackage{graphics}
\usepackage{amssymb}
\usepackage{wrapfig}
\usepackage{enumitem}
\usepackage{placeins}
\usepackage{caption}
\usepackage{xcolor}

\newcommand{\mysubsection}[1]{\medskip\noindent\textbf{#1}}

\usepackage{tikz,ifthen,pgfplots}

\usetikzlibrary{arrows,trees,backgrounds,automata,shapes,decorations,plotmarks,fit,calc,positioning,shadows,chains}
\usepackage{amsmath}

\definecolor{color0}{RGB}{228,87,46}
\definecolor{color1}{RGB}{23,190,187}
\definecolor{color2}{RGB}{255,201,20}
\definecolor{color3}{RGB}{46,40,42}
\definecolor{color4}{RGB}{118,176,65}

\definecolor{color0}{RGB}{250,121,33}
\definecolor{color1}{RGB}{254,153,32}
\definecolor{color2}{RGB}{185,164,76}
\definecolor{color3}{RGB}{86,110,61}
\definecolor{color4}{RGB}{12,71,103}

\definecolor{color0}{RGB}{228,253,225}
\definecolor{color1}{RGB}{138,203,136}
\definecolor{color2}{RGB}{100,131,129}
\definecolor{color3}{RGB}{87,87,97}
\definecolor{color4}{RGB}{255,191,70}

\definecolor{color0}{RGB}{226,59,62}
\definecolor{color1}{RGB}{243,114,44}
\definecolor{color2}{RGB}{248,150,30}
\definecolor{color3}{RGB}{249,199,79}
\definecolor{color4}{RGB}{126,179,86}
\definecolor{color5}{RGB}{67,170,139}
\definecolor{color6}{RGB}{39,125,161}
\definecolor{color7}{RGB}{21,49,60}
\definecolor{color8}{RGB}{180,215,228}

\tikzstyle{every pin edge}=[<-,shorten <=1pt]
\tikzstyle{neuron}=[circle,fill=black!25,minimum size=17pt,inner sep=0pt]
\tikzstyle{input neuron}=[neuron, fill=green!50]
\tikzstyle{output neuron}=[neuron, fill=red!50]
\tikzstyle{hidden neuron}=[neuron, fill=blue!50]
\tikzstyle{annot} = [text width=4em, text centered]
\usetikzlibrary{calc}

\newcommand{\forwardOutput}{\texttt{FORWARD}\xspace}
\newcommand{\leftOutput}{\texttt{LEFT}\xspace}
\newcommand{\rightOutput}{\texttt{RIGHT}\xspace}

\newcommand{\relu}{\text{ReLU}\xspace{}}

\newcommand{\sat}{\texttt{SAT}}
\newcommand{\unsat}{\texttt{UNSAT}}
\newcommand{\timeout}{\texttt{TIMEOUT}}
\newcommand{\memout}{\texttt{MEMOUT}}

\newcommand{\scenarioOne}{\text{\emph{avoid back-and-forth rotation}}}
\newcommand{\scenarioTwo}{\text{\emph{avoid turns larger than 180$^{\circ}$}}}
\newcommand{\scenarioThree}{\text{\emph{avoid turning when clear}}}

\definecolor{nnedgecolor}{RGB}{90,90,90}
\tikzstyle{every pin edge}=[<-,shorten <=1pt]
\tikzstyle{every path}=[draw=color7!50]
\tikzstyle{neuron}=[circle,fill=black!25,minimum size=17pt,inner sep=0pt]
\tikzstyle{input neuron}=[neuron, fill=color4]
\tikzstyle{output neuron}=[neuron, fill=color0]
\tikzstyle{hidden neuron}=[neuron, fill=color6!80]
\tikzstyle{annot} = [text width=4em, text centered]
\tikzstyle{nnedge} = [-{stealth},shorten >=0.1cm, shorten <=0.05cm,line 
width=0.8pt,nnedgecolor]
\usetikzlibrary{calc}

\usepackage{listings}

\DeclareFixedFont{\ttb}{T1}{txtt}{bx}{n}{7.5} 
\DeclareFixedFont{\ttm}{T1}{txtt}{m}{n}{7.5}  

\definecolor{deepblue}{rgb}{0,0,0.5}
\definecolor{deepred}{rgb}{0.6,0,0}
\definecolor{deepgreen}{rgb}{0,0.5,0}

\usepackage{lipsum}
\newcommand\blfootnote[1]{%
  \begingroup
  \renewcommand\thefootnote{}\footnote{#1}%
  \addtocounter{footnote}{-1}%
  \endgroup
}

\title{Constrained Reinforcement Learning for Robotics \\ via Scenario-Based Programming}

\author{
Davide Corsi$^{1, 2, *}$, Raz Yerushalmi$^{1, 3, *}$, 
\\
\textbf{Guy Amir}$^1$, \textbf{Alessandro Farinelli}$^2$, \textbf{David Harel}$^3$ and \textbf{Guy Katz}$^1$\\
\\
$^1$The Hebrew University of Jerusalem \texttt{\{guyam, guykatz\}@cs.huji.ac.il} \\ 
$^2$University of Verona \texttt{\{davide.corsi, alessandro.farinelli\}@univr.it} \\
$^3$The Weizmann Institute of Science \texttt{\{raz.yerushalmi, david.harel\}@weizmann.ac.il} 
}

\begin{document}
\maketitle

\blfootnote{$^*$ Both authors contributed equally.}

\begin{abstract}
  Deep reinforcement learning (DRL) has achieved groundbreaking
  successes in a wide variety of 
  robotic applications.
  A natural consequence is the adoption of this paradigm for safety-critical tasks, where human safety and expensive hardware can be involved.
  In this context, it is crucial to optimize the performance of DRL-based agents while providing guarantees about their behavior.
  This paper presents a novel technique for incorporating domain-expert knowledge into a \textit{constrained DRL} training loop. Our technique exploits the \textit{scenario-based programming} paradigm, which is designed to allow specifying such knowledge in a simple and intuitive way. 
  We validated our method on the popular robotic mapless navigation problem, in simulation, and on the actual platform.
  Our experiments demonstrate that using our approach to leverage expert knowledge dramatically improves the safety and the performance of the agent.
\end{abstract}



\section{Introduction}
\label{sec:introduction}

In recent years, \emph{deep neural networks} (DNNs) have achieved
state-of-the-art results in a large variety of tasks, including image
recognition~\citep{Du18}, game playing~\citep{MnKaSi13}, 
protein folding~\citep{JuEvPr21}, and more. In particular, \emph{deep
  reinforcement learning} (DRL)~\citep{SuBa18} has emerged as a
popular paradigm for training DNNs that perform complex tasks through
 continuous interaction with their environment. Indeed, DRL models
have proven remarkably useful in robotic control tasks, such as
navigation~\citep{KuDeDe19} and manipulation~\citep{NgLa19, CoMaFa21},
where they often outperform classical algorithms~\citep{ZhZh21}.  The
success of DRL-based systems has naturally led to their integration as control policies in safety-critical tasks, such as
autonomous driving~\citep{SaAbPe17}, surgical
assistance~\citep{PoCoMa21}, flight control~\citep{KoMaWe19}, and
more. Consequently, the learning community has been seeking to
create DRL-based controllers that simultaneously demonstrate high
\emph{performance} and high \emph{reliability}; i.e., are able to
perform their primary tasks while adhering to some prescribed
properties, such as safety and
robustness.

An emerging family of approaches for achieving these two
goals, known as \emph{constrained DRL}~\citep{AcHeTa17}, attempts to
simultaneously optimize two functions: the \textit{reward}, which
encodes the main objective of the task; and the \textit{cost}, which
represents the safety constraints.  Current state-of-the-art
algorithms include IPO~\citep{LiDiLi20}, SOS~\citep{MaCoFa21b},
CPO~\citep{AcHeTa17}, and Lagrangian
approaches~\citep{RaAcAm19}. Despite their success in some
applications, these methods generally suffer from significant
setbacks: (i) there is no uniform and human-readable way of defining
the required 
safety constraints; (ii) it is unclear how to encode
these constraints as a signal for the training algorithm; and (iii)
there is no clear method for balancing cost and reward during
training, and thus there is a risk of producing sub-optimal policies.

In this paper, we present a novel approach for addressing these
challenges, by enabling users to encode constraints into the DRL
training loop in a simple yet powerful way.  Our approach 
generates policies that strictly adhere to these user-defined
constraints without compromising performance.  

We achieve this by
extending and integrating two approaches: the \textit{Lagrangian-PPO}
algorithm~\citep{RaAcAm19} 
for DRL training, and the \emph{scenario-based programming} (SBP)~\citep{DaHa01, HaMaWe12ACM}
framework for encoding user-defined constraints. 

Scenario-based programming is a software engineering paradigm 
intended to allow engineers to create a complex system in a way that
is aligned with how humans perceive that system.
A scenario-based program is comprised of scenarios, each of which
describes a single desirable (or undesirable) behavior 
of the system at hand; and these scenarios are then combined to run
simultaneously, in order to produce cohesive system behavior.  We show
how such scenarios can be used to directly incorporate
subject-matter-expert (SME) knowledge into the training process, thus
forcing the resulting agent's behavior to abide various safety,
efficiency and predictability requirements.

In order to demonstrate the usefulness of our approach to robotic
systems, we used it to train a policy for performing \emph{mapless
  navigation}~\citep{ZhSpBo17, TaPaLi17} by the
Robotis Turtlebot3 platform. 

While common
DRL-training techniques were shown to give rise to high-performance policies for
this task~\citep{MaFa20}, these policies are often 
unsafe, inefficient, or unpredictable, thus dramatically limiting
their  potential deployment in real-world
systems~\citep{MaCoFa21a, MaCoFa21b}.  

Our experiments demonstrate
that, by using our novel approach and injecting subject-matter expert
knowledge into the training process, we are able to generate
trustworthy policies that are both safe and high performance.
To have a complete assessment of the resulting behaviors, we performed a formal verification analysis~\citep{KaBaDiJuKo17, LiArLaBaKo19} of various predefined safety properties that proved that our approach generates safe agents to deploy in \emph{any} environment. 

\section{Background}
\label{sec:background}

\mysubsection{Deep Reinforcement Learning.}
Deep reinforcement learning~\citep{Li17} is a specific paradigm for training deep neural networks~\citep{GoBeCo16}.
In DRL, the training objective is to find a \emph{policy} that maximizes the \textit{expected cumulative discounted reward} $R_t=\mathbb{E}\big[\sum_{t}\gamma^{t}\cdot r_t\big]$, where $\gamma \in \big[0,1\big]$ is the \textit{discount factor}, a hyperparameter that controls the impact of past decisions on the total expected reward. 
The \emph{policy}, denoted as $\pi_\theta$, is a probability
distribution that depends on the parameters $\theta$ of the DNN, 
which maps an observed \emph{environment state} $s$ to an \emph{action} $a$. 
Proximal policy optimization (PPO) is a state-of-the-art DRL algorithm
for producing $\pi_\theta$~\citep{ShWoDh17}. A key characteristic of
PPO is that it limits the gradient step size between two consecutive policy updates
during training, to avoid changes that can drastically modify
$\pi_\theta$~\citep{ScLeAbJo15}.

In mission-critical tasks, the concept of optimality often goes beyond
the maximization of a reward, and also involves ``hard'' safety
constraints that the agent must respect.  A \textit{constrained markov
  decision process} (CMDP) is an alternative framework for sequential
decision making, which includes an additional signal: the \textit{cost
  function}, defined as
$C: \mathcal{S} \times \mathcal{A} \rightarrow \mathbb{R}$, whose
expected values must remain below a given threshold $d \in
\mathbb{R}$. CMDP can support multiple cost functions and their 
thresholds, denoted by $\{C_k\}$ and $\{d_k\}$, respectively.
The set of \textit{valid} policies for a CMDP is defined as:
\begin{equation}
    \Pi_\mathcal{C} := \{\pi_\theta \in \Pi : \ \forall k, \ J_{C_k}(\pi_\theta) \leq d_k\}
\label{eq:background:valid_cmdp}
\end{equation}
\noindent where $J_{C_k}(\pi_\theta)$ is the expected sum of the
$k^{th}$ cost function over the trajectory and $d_k$ is the
corresponding threshold. Intuitively, the objective is to find a
policy function that respects the constraints (i.e., is
\textit{valid}) and which also maximizes the expected reward (i.e., is \textit{optimal}). A natural way to encode constraints in a classical optimization problem is by using \textit{Lagrange multipliers}. Specifically, in DRL, a possible approach is to transform the constrained problem into the corresponding dual unconstrained version~\citep{LiDiLi20, AcHeTa17}. The optimization problem can then be encoded as follows:
\begin{equation}
   J(\theta) = \min_{\pi_\theta} \max_{\lambda \geq 0} \mathcal{L}(\pi_\theta, \lambda) = \min_{\pi_\theta} \max_{\lambda \geq 0} J_R(\pi_\theta) - \sum_K \lambda_k(J_{C_k}(\pi_\theta) - d_k)
\label{eq:background:lagrangian}
\end{equation}
Crucially, the optimization of the function $J(\theta)$ can be carried
out by applying any \textit{policy gradient} algorithm, a common
implementation is based on PPO~\citep{RaAcAm19}.


\mysubsection{Scenario-Based Programming.}
Scenario-based programming (SBP)~\citep{DaHa01,HaMa03} is a paradigm
designed to facilitate the development of reactive systems, by
allowing engineers to program a system in a way that is close to how
it is perceived by humans --- with a focus on inter-object,
system-wide behaviors.  In SBP, a system is composed of
\emph{scenarios}, each describing a single, desired or undesired
behavioral aspect of the system; and these scenarios are then executed
in unison as a cohesive system.

 An execution of a scenario-based (SB) program is 
 formalized as a discrete sequence of events.
 At each time-step, the scenarios synchronize with each other to
 determine the next event to be triggered.
 Each scenario declares events that it \emph{requests} and events that it
\emph{blocks}, corresponding to desirable and undesirable (forbidden) behaviors
from its perspective; and also events that it passively \emph{waits-for}.
After making these declarations, 
the scenarios are temporarily suspended, and an \emph{event-selection
  mechanism} triggers a single event that was requested by at least
one scenario and blocked by none.  Scenarios that requested or waited
for the triggered event wake up, perform local actions, and then
synchronize again; and the process is repeated ad infinitum.  The
resulting execution thus complies with the requirements and
constraints of each of the individual
scenarios~\citep{HaMa03,HaMaWe12ACM}. For a formal definition of SBP,
see the paper from ~\citet{HaMaWe12ACM}.

Although SBP is implemented in many high-level languages, it is often
convenient to think of scenarios as transition systems, where each
state corresponds to a synchronization point, and each edge corresponds
to an event that could be triggered.

\begin{figure*}[hb]
	\centering
	\includegraphics[width=0.9\textwidth]{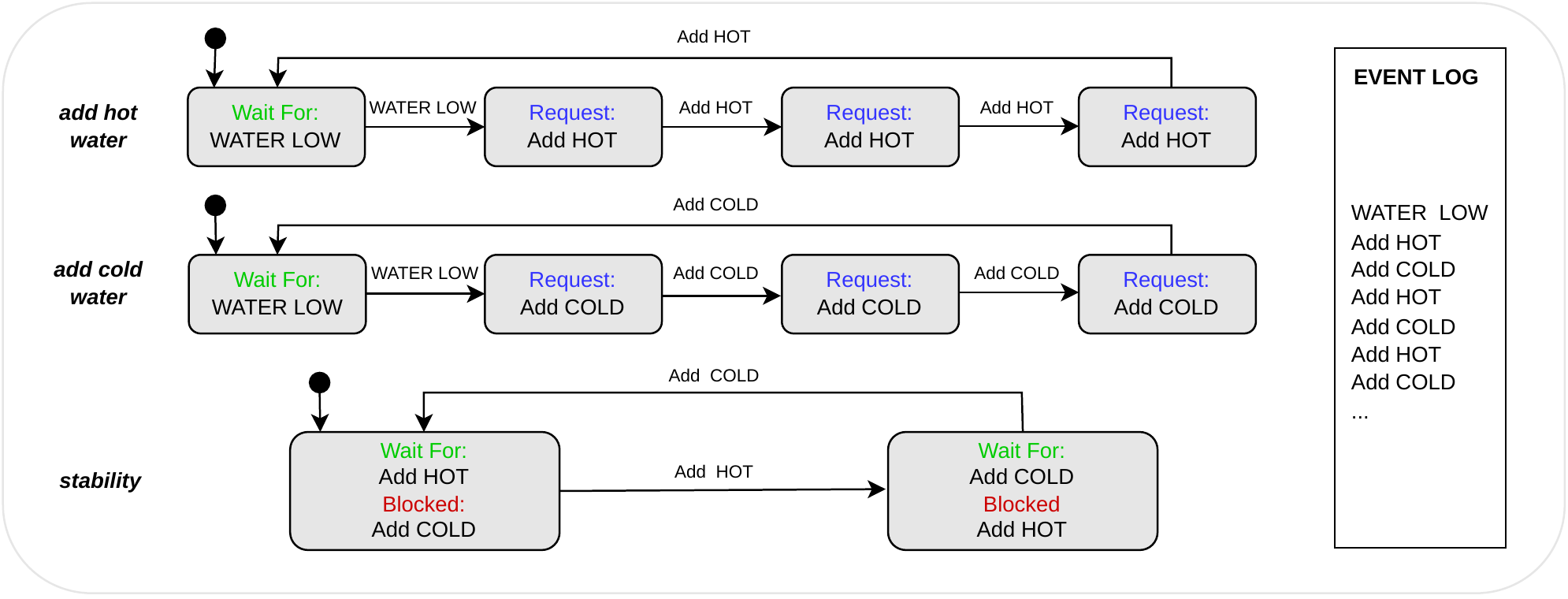}
	\caption{A scenario-based program for controlling a water
          tank. The small black circle indicates the initial state.
          The example is inspired by the work of~\citet{HaKaMaWe12}.
        }
	\label{fig:watertap}
\end{figure*}

Fig.~\ref{fig:watertap} uses that representation to depict 
a simple SB program that controls the temperature and water-level in a
water tank (borrowed from~\citep{HaKaMaWe12}). 
The scenarios \emph{add hot water} and \emph{add cold water} repeatedly wait for \texttt{WATER LOW}
events, and then request three times the event \texttt{Add HOT} or
\texttt{Add COLD}, respectively.  Since these six events may be
triggered in any order by the event selection mechanism, new
scenario \emph{stability} is added to keep the 
water temperature stable, achieved by alternately blocking 
\texttt{Add HOT} and \texttt{Add COLD} events.

The resulting execution trace is shown in the event log.

A particularly useful trait of SBP is that the resulting models are
amenable to model checking, and facilitate compositional
verification~\citep{HaLaMaWe11BPMC,HaKaMaWe15,KaBaHa15,Ka13,HaKaLaMaWe15}. Thus,
it is often possible to apply formal verification to ensure that a
scenario-based model satisfies various criteria, either as a
stand-alone model or as a component within a larger system. Automated
analysis techniques can also be used to execute scenario-based models
in distributed
architectures~\citep{HaKaKaMaWeWi15,StGrGrHaKaMa18,StGrGrHaKaMa17,GrGrKaMaGlGuKo16,HaKaKa13},
to automatically repair these
models~\citep{HaKaMaWe14,HaKaMaWe12,Ka21}, and to augment them in
various ways, e.g., as part of the Wise Computing
initiative~\citep{HaKaMaMa18,HaKaMaMa16,HaKaMaMa16b}.
In our context, SBP is an attractive choice for the incorporation of
domain-specific knowledge into a DRL agent training process, due to
being formal, fully executable and support of incremental
development~\citep{GoMaMe12Spaghetti,AlArGoHa14CognitiveLoad,KaEl21,Ka21b,Ka20}.
Moreover, the language it uses enables domain-specific experts to
directly express their requirements specifications as an SB program.

\mysubsection{Formal Verification of DNNs and DRL.}
A DNN verification algorithm receives the following
inputs~\citep{KaBaDiJuKo17}: a trained DNN $N$, a precondition $P$ on
the DNN's inputs, and a postcondition $Q$ on $N$'s output.  The
precondition is used to limit the input assignments to inputs of
interest, or to express some assumption the user has regarding the
environment (e.g., that an image-recognition DNN will only be
presented with certain pixel values).  The postcondition typically
encodes the \textit{negation} of the behavior we would like $N$ to
exhibit on inputs that satisfy $P$. Then, the verification algorithm
searches for an input $x'$ that satisfies the given conditions (i.e.,
$P(x') \wedge Q(N(x'))$), and returns exactly one of the following
outputs:
\begin{inparaenum}[(i)]
\item \sat, indicating the query is satisfiable. Due to the
  postcondition $Q$ encoding the negation of the required property,
  this result indicates that the wanted property is violated in some
  cases.  Modern verification engines also supply a concrete input
  $x'$ that satisfies the query, and hence, a valid input that
  triggers a bug, such as an incorrect classification; or
\item \unsat, indicating that there does not exist such an $x'$, and
  thus --- that the desired property always holds.
\end{inparaenum}

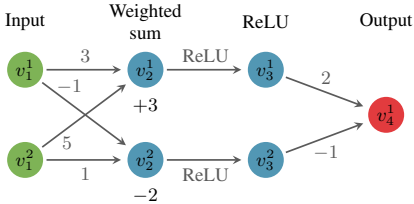
\begin{figure}[htp]
	\begin{center}
		\scalebox{0.8} {
			\def\layersep{2.0cm}
			\begin{tikzpicture}[shorten >=1pt,->,draw=black!50, node 
				distance=\layersep,font=\footnotesize]
				
				\node[input neuron] (I-1) at (0,-1) {$v^1_1$};
				\node[input neuron] (I-2) at (0,-2.5) {$v^2_1$};
				
				\node[hidden neuron] (H-1) at (\layersep,-1) {$v^1_2$};
				\node[hidden neuron] (H-2) at (\layersep,-2.5) {$v^2_2$};
				
				\node[hidden neuron] (H-3) at (2*\layersep,-1) {$v^1_3$};
				\node[hidden neuron] (H-4) at (2*\layersep,-2.5) {$v^2_3$};
				
				\node[output neuron] at (3*\layersep, -1.75) (O-1) {$v^1_4$};
				
				\draw[nnedge] (I-1) --node[above] {$3$} (H-1);
				\draw[nnedge] (I-1) --node[above, pos=0.3] {$\ -1$} (H-2);
				\draw[nnedge] (I-2) --node[below, pos=0.3] {$5$} (H-1);
				\draw[nnedge] (I-2) --node[below] {$1$} (H-2);
				
				\draw[nnedge] (H-1) --node[above] {$\relu$} (H-3);
				\draw[nnedge] (H-2) --node[below] {$\relu$} (H-4);
				
				\draw[nnedge] (H-3) --node[above] {$2$} (O-1);
				\draw[nnedge] (H-4) --node[below] {$-1$} (O-1);

				\node[below=0.05cm of H-1] (b1) {$+3$};
				\node[below=0.05cm of H-2] (b2) {$-2$};
				
				\node[annot,above of=H-1, node distance=0.8cm] (hl1) {Weighted 
					sum};
				\node[annot,above of=H-3, node distance=0.8cm] (hl2) {ReLU };
				\node[annot,left of=hl1] {Input };
				\node[annot,right of=hl2] {Output };
			\end{tikzpicture}
		}
		\captionsetup{size=small}
		\captionof{figure}{A toy DNN.}
		\label{fig:toyDnn}
	\end{center}
\end{figure}

For example, suppose we wish to guarantee that for all
non-negative inputs $x=\langle v_1^1,v_1^2\rangle$, the DNN in
Fig.~\ref{fig:toyDnn} always outputs a value strictly smaller than
$40$; i.e., that that $N(x)=v_4^1 < 40$. This property can be encoded
as a verification query consisting of a precondition that restrict the
inputs to the desired range, i.e.,
$P=(v_1^1 \geq 0) \wedge (v_1^2 \geq 0)$, and by setting
$Q=(v_4^1\geq 40)$, which is the \textit{negation} of the desired
property.  In this case, a sound verifier will return \sat, alongside
a feasible counterexample such as $x=\langle 2, 3\rangle$, which
produces the output $v_4^1=48 \geq\ 40$ when fed to the DNN. Hence,
the property does not always hold.

Originally, DNN verification engines were designed the verify the
correct behaviour of feed-forward DNNs~\citep{ KaBaDiJuKo17,
  GeMiDrTsChVe18, WaPeWhYaJa18, LyKoKoWoLiDa20, HuKwWaWu17}. However,
in recent years, the verification community has also designed
verification methods tailored for DRL systems~\citep{CoMaFa21,
  BaGiPa21, ElKaKaSc21, AmScKa21, AmCoYeMaHaFaKa22}. These methods
include techniques for encoding multiple invocations of the agent in
question, when interacting with a reactive environment over
multiple time-steps.

\begin{figure*}[b]
	\centering
	\includegraphics[width=0.7\textwidth]{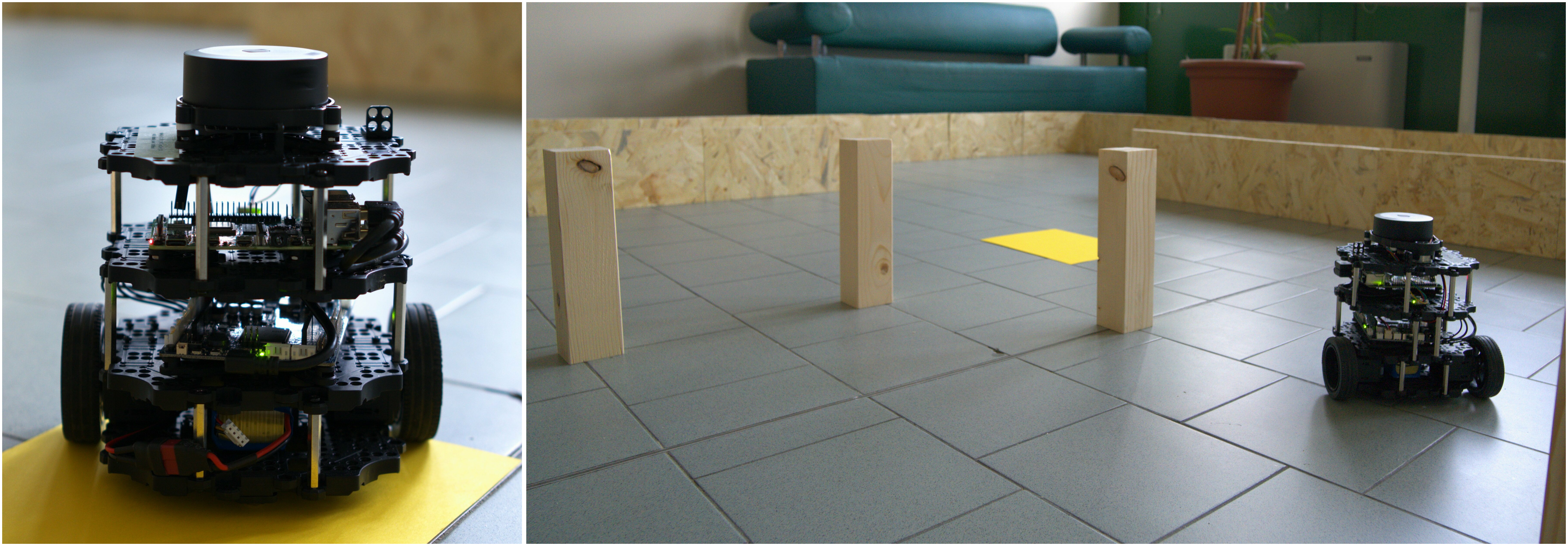}
	\caption{The Robotis Turtlebot3 platform.}
	\label{fig:intro:robot}
\end{figure*} 

\section{Expressing DRL Constraints using Scenarios}
\label{sec:methods}
\mysubsection{Mapless Navigation.}
We explain and demonstrate our proposed technique using the
\emph{mapless navigation} problem, in which a robot is required to
reach a given target efficiently while avoiding collision with
obstacles. Unlike in classical planning, the
robot is not given a map of its surrounding environment and can rely
only on local observations --- e.g., from lidar sensors or cameras.
Thus, a successful agent needs to be able to adjust its
strategy dynamically, as it progresses towards its target.  Mapless navigation has
been studied extensively and is considered difficult to
solve. Specifically, the local nature of the problem renders learning
a successful policy extremely challenging and hard to solve using
classical algorithms~\citep{PfShTu18}. Prior work has shown DRL
approaches to be among the most successful for tackling this task,
often outperforming hand-crafted algorithms~\citep{MaFa20}.

As a platform for our study, we used the \textit{Robotis
  Turtlebot 3} platform (Turtlebot, for short; see
Fig.~\ref{fig:intro:robot}), which is widely used in the
community~\citep{NaShVa21, AmSl19}. The Turtlebot is capable of
horizontal navigation and is equipped with lidar sensors for
detecting nearby obstacles. In order to train DRL policies for
controlling the Turtlebot, we built a simulator based on the
\textit{Unity3D} engine~\citep{JuBeTe18}, which is compatible with the
\textit{Robotic Operating System} (ROS)~\citep{QuCoGe09} and allows a fast transfer to the actual platform
(\textit{sim-to-real}~\citep{ZhQuWe20}).

We used a hybrid  reward function, which
includes a discrete component for the
terminal states (``collision'', or ``reached target''), and a
continuous component for the
non-terminal states. Formally:
\begin{equation}
    R_t = \begin{cases}
    \pm 1 \hfill \text{terminal states} \\
    (dist_{t-1} - dist_{t}) \cdot \alpha - \beta \hspace{30px} \text{otherwise}\\
    \end{cases}
\label{eq:methods:reward_function}
\end{equation}
Where $dist_k$ is the distance from the target at time $k$; $\alpha$ is a
normalization factor; and $\beta$ is a penalty, intended to encourage
the robot to reach the target quickly (in our experiments, we
empirically set $\alpha=3$ and $\beta=0.001$). 
Additionally, in terminal states, we increase the reward by $1$ if the
target is reached, or decrease it by $1$ in case of collision. 

For our DNN topology, we used an 
architecture that was shown to be successful 
in a similar setting~\citep{MaFa20}: (i) an input layer of
nine neurons, including seven for the lidar scans and two for the
polar coordinates of the target; (ii) two fully-connected hidden
layers of 32 neurons each; and (iii) an output layer of three neurons
for the discrete actions (i.e., move \forwardOutput, turn \leftOutput, and turn \rightOutput). 

In Section \ref{sec:methods:lagrangian}, we provide details about the
training algorithm we used.
Using the reward defined in Eq.~\ref{eq:methods:reward_function}, we
were able to train agents that achieved high performance
--- i.e., obtained a success rate of approximately 95\%, where
``success'' means that the robot reached its target
without colliding into walls or obstacles.

Analyzing the trained agents further, we observed that even DRL agents
that achieved a high success rate may demonstrate highly undesirable
behavior in different scenarios.  One such behavior is a sequence of
back-and-forth turns, that causes the robot to waste time and energy. Another undesirable behavior
is when the agent makes a lengthy sequence of right turns 
instead of a much shorter sequence of left turns (or vice versa),
 wasting time and energy. A third undesirable behavior that we
observed is that the agent might decide not to move forward towards a
target that is directly ahead, even when the path is clear. Our goal
was thus to use our approach to remove these undesirable behaviors.


\mysubsection{A Rule-Based Approach.}
Following the approach of~\citep{Yerushalmi2022ScenarioAssistedDR}, we 
integrated a scenario-based program into the DRL training
process, in order to remove the aforementioned undesirable behaviors.
More concretely, we created specific scenarios to rule out each of the
three aforementioned undesirable behaviors we observed.  

To accomplish this,
we created a mapping between each possible action
$a_t\in \{\text{Move \forwardOutput, Turn \leftOutput, Turn \rightOutput}\}$ of the DRL
agent and a dedicated event
$e_{a_t}\in \{\text{SBP\_MoveForward, SBP\_TurnLeft,
  SBP\_TurnRight}\}$ within the scenario-based program. These events
allow the various scenarios to keep track and react to the agent's
actions.  Similarly to \citep{Yerushalmi2022ScenarioAssistedDR}, we
refer to these $e_{a_t}$ events as \emph{external events}, indicating
that they can only be triggered when requested from outside the SB
program proper. 

By convention, we assume that after each triggering of
a single, external event, the scenario-based program executes a
sequence of internal events (a
\emph{super-step}~\citep{Yerushalmi2022ScenarioAssistedDR}), until
it returns to a steady-state and then waits for another external event.

The novelty of our approach, compared to 
\citep{Yerushalmi2022ScenarioAssistedDR}, is in the strategy by which we use scenarios to affect
the training process. 
Specifically, we define the DRL cost function to correspond to 
violations of scenario constraints by the DRL agent. Whenever the
agent selects an action that is mapped to a \emph{blocked} SBP event, we increase the \emph{cost}.
This approach is described further in Section~\ref{sec:methods:lagrangian},
and constitutes a general 
method for injecting explicit constraints
(expressed, e.g., by scenarios) directly into the policy
optimization process.

\begin{figure*}[hb]
	\centering
  \scalebox{0.90} {
    \subfigure[\scenarioOne]{\includegraphics[width=0.32\textwidth]{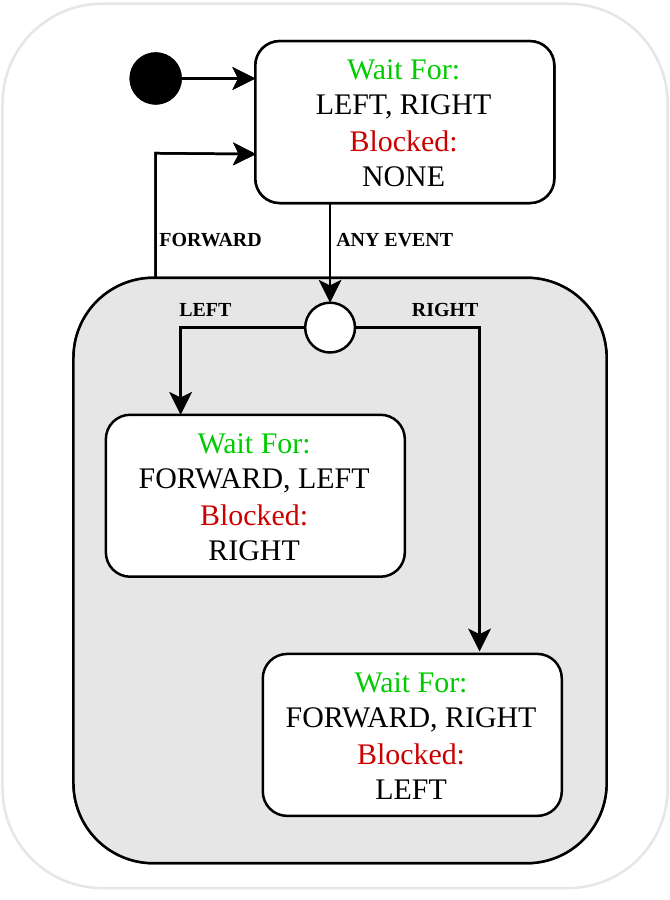}
    \label{fig:methods:all_scenarios_a}}
    \hspace{20px}
    \subfigure[\scenarioTwo]
    {\includegraphics[width=0.32\textwidth]{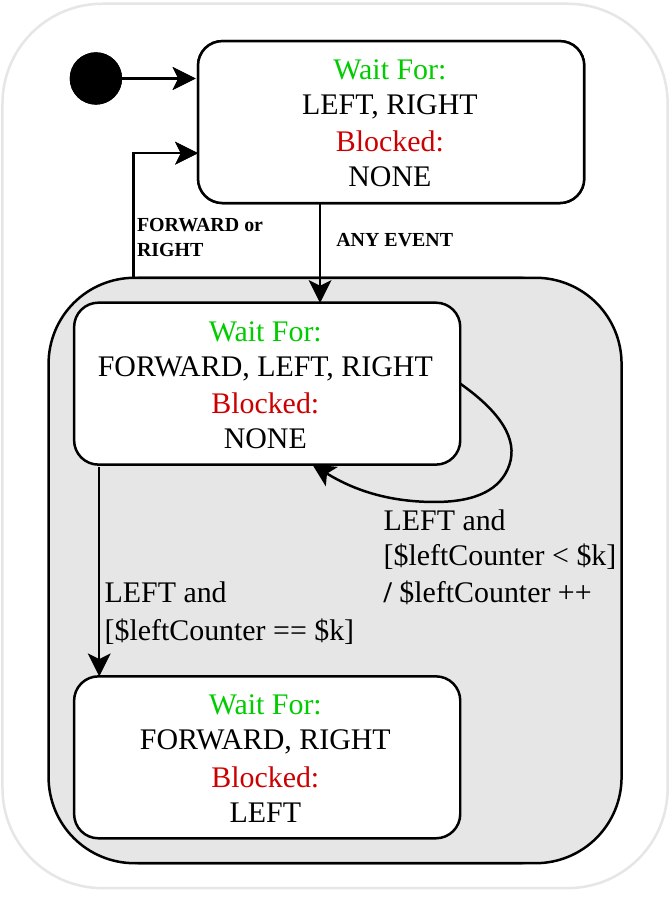}
        \label{fig:methods:all_scenarios_b}}
    \hspace{20px}
\subfigure[\scenarioThree]
        {\includegraphics[width=0.32\textwidth]{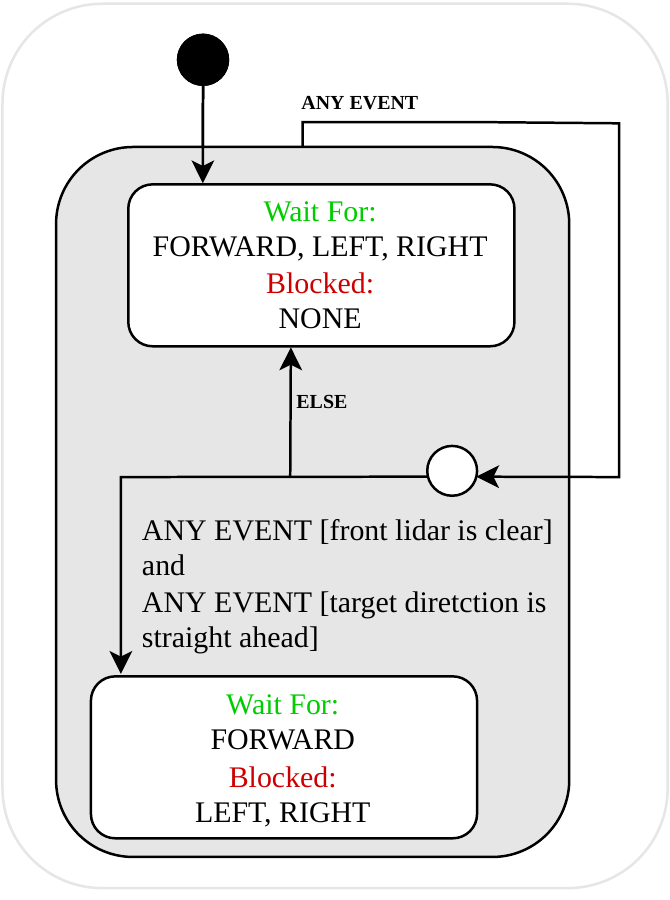}
    \label{fig:methods:all_scenarios_c}}
	}
	\caption{A visualization of the three scenarios. 
	Figure (b) refers to the \textit{Left turns part}~only.
	``Wait For'' and ``Blocked'' in the state-blob indicates events that the scenario waits for or blocks, respectively. The events \texttt{SBP\_MoveForward, SBP\_TurnLeft and SBP\_TurnRight} are represented respectively, by {FORWARD, LEFT, RIGHT}.}
	\label{fig:methods:all_scenarios}
\end{figure*}

\mysubsection{Example: Constraint Scenarios.}
Considering again our Turtlebot mapless navigation case study, we
created scenarios for discouraging the three undesirable behaviors 
we had previously observed. 
The scenarios are visualized in
Fig.~\ref{fig:methods:all_scenarios}, using an amalgamation of
Statecharts and SBP graphical notation
languages~\citep{harel1987statecharts, marron2018embedding}.

  Scenario \scenarioOne{} (Fig.~\ref{fig:methods:all_scenarios_a})
  seeks to prevent in-place, back-and-forth turns by the robot, to conserve
  time and energy.

  Scenario \scenarioTwo{} (Fig.~\ref{fig:methods:all_scenarios_b})
  seeks to prevent left turns in angles that are greater than
  $180^{\circ}$, to conserve time and energy (the right-turn
  case is symmetrical). A forward slash indicates an action that is performed when
    a transition is taken; square brackets denote guard conditions, 
    and \$k and \$leftCounter are variables.
    Each turn rotates the robot by $30^{\circ}$, and so we set
    $k=7$.

    Scenario \scenarioThree{}
    (Fig.~\ref{fig:methods:all_scenarios_c}) seeks to force the agent
    to move towards the target when it is ahead, and there is a clear
    path to it. This is performed by blocking any turn actions when
    this situation occurs. 
    Triggered events carry data, which can be referenced by guard conditions.  

\begin{figure}[ht]
\begin{lstlisting}[language=Python]
def SBP_avoidBackAndForthRotation():
    blockedEvList = []
    waitforEvList = [BEvent("SBP_MoveForward"),
                     BEvent("SBP_TurnLeft"),
                     BEvent("SBP_TurnRight")]
    while True:
        lastEv = yield {waitFor: waitforEvList, block: blockedEvList}
        if lastEv != BEvent("SBP_TurnLeft")  
            and lastEv != BEvent("SBP_TurnRight"):
            blockedEvList = []
        else:
            blocked_ev = BEvent("SBP_TurnRight") 
                if lastEv == BEvent("SBP_TurnLeft")
                else BEvent("SBP_TurnLeft")
            # Blocking!
            blockedEvList.append(blocked_ev)
\end{lstlisting}
  \caption{The Python implementation of scenario~\scenarioOne. The
    code waits for any of the possible events:
    \emph{SBP\_MoveForward, SBP\_TurnLeft} and
    \emph{SBP\_TurnRight}. Upon receiving \emph{SBP\_TurnLeft}, it
    blocks \emph{SBP\_TurnRight}, and upon receiving
    \emph{SBP\_TurnRight}, it blocks \emph{SBP\_TurnLeft}.  Upon
    receiving \emph{SBP\_MoveForward}, it clears any blocking.}
\label{fig:methods:scenario-1}
\end{figure}
The Python implementation of \scenarioOne{} is presented in
Fig.~\ref{fig:methods:scenario-1}, while a complete listing of all three scenarios appears in Appendix ~\ref{sec:appendix:sbm}.

\section{Using Scenarios in DRL Training}
\label{sec:methods:lagrangian}
Even after defining constraints as an SB program, obtaining a
differentiable function for the training process is not straightforward.
 We propose to use the following binary (indicator) function to this end:
\begin{equation*}
   c_k(s_t, a, s_{t+1}) = I(\text{the tuple } \langle s_t, a, s_{t+1} \rangle \text{ is a blocked state in the SB program, by the $k^{th}$ rule}) 
\end{equation*}
Intuitively, summing the values of $c_k$ over a training episode yields
the number of violations to the $k^{th}$ scenario rule during a single
trajectory.
This value can be treated as a cost function, the corresponding objective function defined as follows:
$J_{C_k}=\sum_I c(s_i, a_i, s_{i+1})$, for a trajectory of $I$ steps.
This value is dependent on the action policy $a$ and is therefore
differentiable on the parameters $\theta$ of the policy through the
\textit{policy gradient theorem}.

\mysubsection{Optimized Lagrangian-PPO.}
In Section~\ref{sec:background} we proposed to relax the 
Lagrangian constrained optimization problem into an unconstrained, 
\textit{min-max} version thereof. 
Taking the gradient of Equation~\ref{eq:background:lagrangian}, and some algebraic manipulation,
we derive the following two simultaneous problems:
\begin{equation}
    \nabla_\theta \mathcal{L}(\pi, \lambda) = \nabla_\theta (J_R(\pi) - \sum_K \lambda_k J_{C_k}(\pi)) \hspace{30px} \forall k , \hspace{10px}  \nabla_{\lambda_k} \mathcal{L}(\pi, \lambda) = -(J_{C_k}(\pi) - d_k)
\label{eq:methods:lagrangian_derivation}
\end{equation}
In closed form, the Lagrangian dual problem would produce exact results. However, when applied using a numerical method like \textit{gradient descent}, it has shown strong instability and the proclivity to optimize only the cost, limiting the exploration and resulting in a poorly-performing agent~\citep{AcHeTa17}. To overcome these problems, we introduce 
three key optimizations that proved crucial to obtaining the results we present in the next section.

\begin{enumerate}
  \item
\emph{Reward Multiplier}: the standard update rule for the policy in a Lagrangian method is given in Equation \ref{eq:methods:lagrangian_derivation}. 
However, as mentioned above, it often fails to  maximize the reward.  
To overcome this failure, we introduce a new parameter $\alpha$, which
we term \textit{reward multiplier}, such that $\alpha \geq \sum_K
\lambda_k$. This parameter is used as a multiplier for the reward
objective:
\begin{equation}
\nabla_\theta \mathcal{L}(\pi, \lambda) =  \nabla_\theta (\alpha \cdot J_R(\pi) - \sum_K \lambda_k J_{C_k}(\pi))
\label{eq:methods:reward_multiplier}
\end{equation}
\item
\emph{Lambda Bounds and Normalization}:
Theoretically, the only constraint on the Lagrangian multipliers is
that they be non-negative. However, when solving numerically, the value of $\lambda_k$
can increase quickly during the early stages of the training, causing the optimizer to focus primarily on the cost functions (Eq. \ref{eq:methods:lagrangian_derivation}), potentially not pushing the policy towards a high performance reward-wise. 
To overcome this, we introduced  dynamic constraints on the multipliers (including the reward multiplier $\alpha$), such that $\sum_K \lambda_k + \alpha = 1$. In order to also enforce the previously mentioned upper bound for $\alpha$, we clipped the values of the multipliers such that $\sum_K \lambda_k \leq \frac{1}{2}$. 
Formally, we perform the following normalization over all the multipliers:
\begin{equation}
    \forall k, \hspace{5px} \lambda_k = \frac{\tilde{\lambda}_{k}}{2(\sum_K \tilde{\lambda}_{k})}  \hspace{40px} \alpha = 1 - \sum_K \lambda_k
\label{eq:methods:lambda_normalization}
\end{equation}
\item
\emph{Algorithmic Implementation}: the primary objective of the previously introduced optimizations is to balance the learning between the reward and the constraints. To further stabilize the training, we introduce additional, minor improvements to the algorithm: (i) \textit{lambda initialization:} we initialize all the Lagrangian multipliers with zero to guarantee a focus on the reward optimization during the early stages of the training (consequently, following Eq. \ref{eq:methods:lambda_normalization}, ~$\alpha=1$); (ii) \textit{lambda learning rate:} to guarantee a smoother update of the Lagrangian multipliers, we scale this parameter to 10\% of the learning rate used for the policy update; and (iii) \textit{delayed start:} we enable the update of the multipliers only when the success rate is above 60\% during the last 100 episodes. Intuitively, this delays the optimization of the cost functions until a minimum performance threshold is reached.
\end{enumerate}

\section{Evaluation}
\label{sec:results}
\mysubsection{Setup.}  We performed training on a distributed cluster
of HP EliteDesk machines, running at 3.00 GHz, with 32 GB RAM.  We
collected data from more than $100$ seeds for each algorithm,
reporting the mean and standard deviation for each learning curve,
following the guidelines of~\citet{CoSiOu19}.

For training purposes,
we built a realistic simulator based on the Unity3D
engine~\citep{JuBeTe18}. Next, we evaluated the
performance of the trained models using a physical Robotis Turtlebot3
robot (Fig.~\ref{fig:intro:robot}) and confirmed that it 
behaved similarly to the behavior observed in our simulations.

\begin{figure*}[htb]
	\centering
  \scalebox{1.0} {
    \subfigure[success rate]{\includegraphics[width=0.33\textwidth]{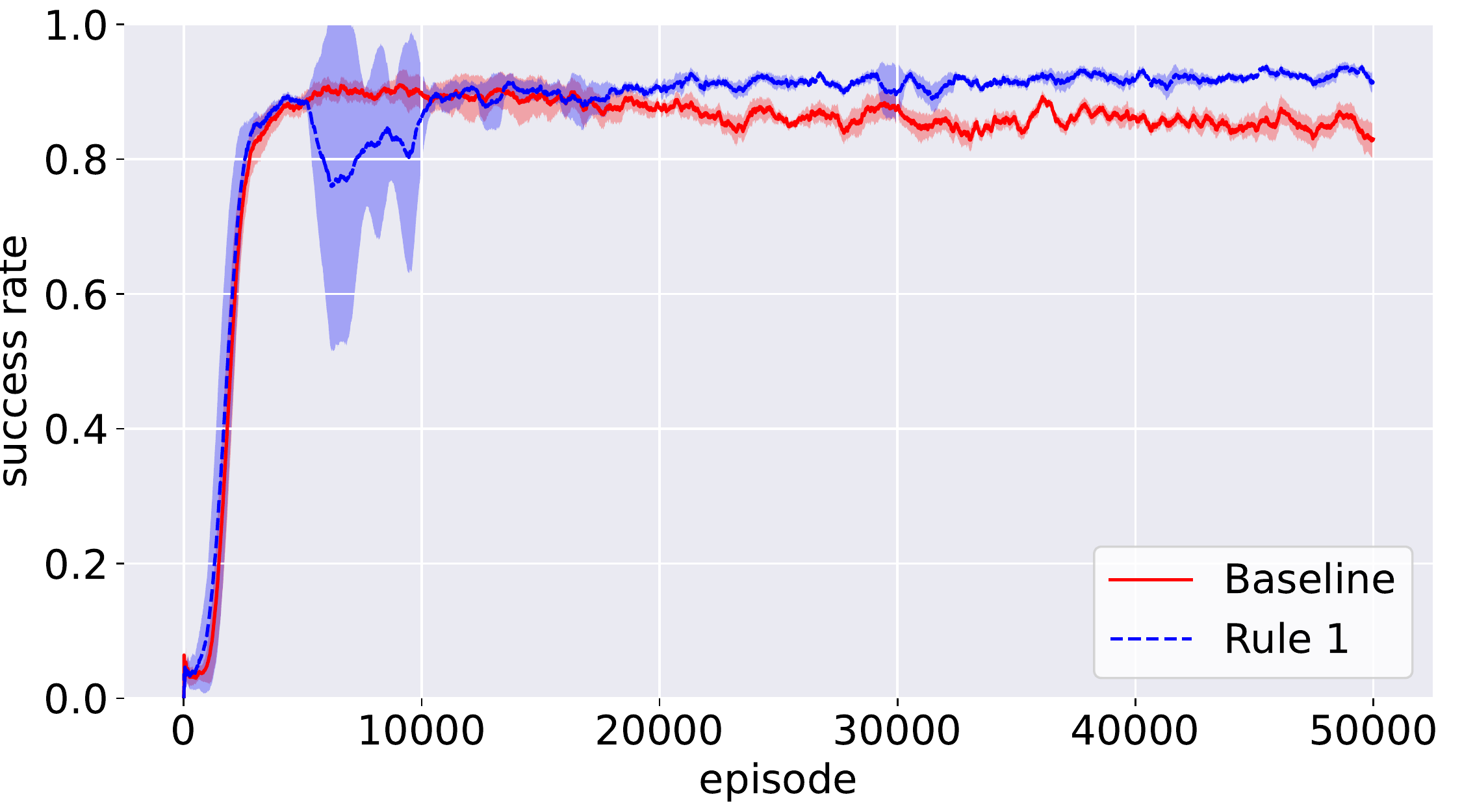}}
    \subfigure[success rate]{\includegraphics[width=0.32\textwidth]{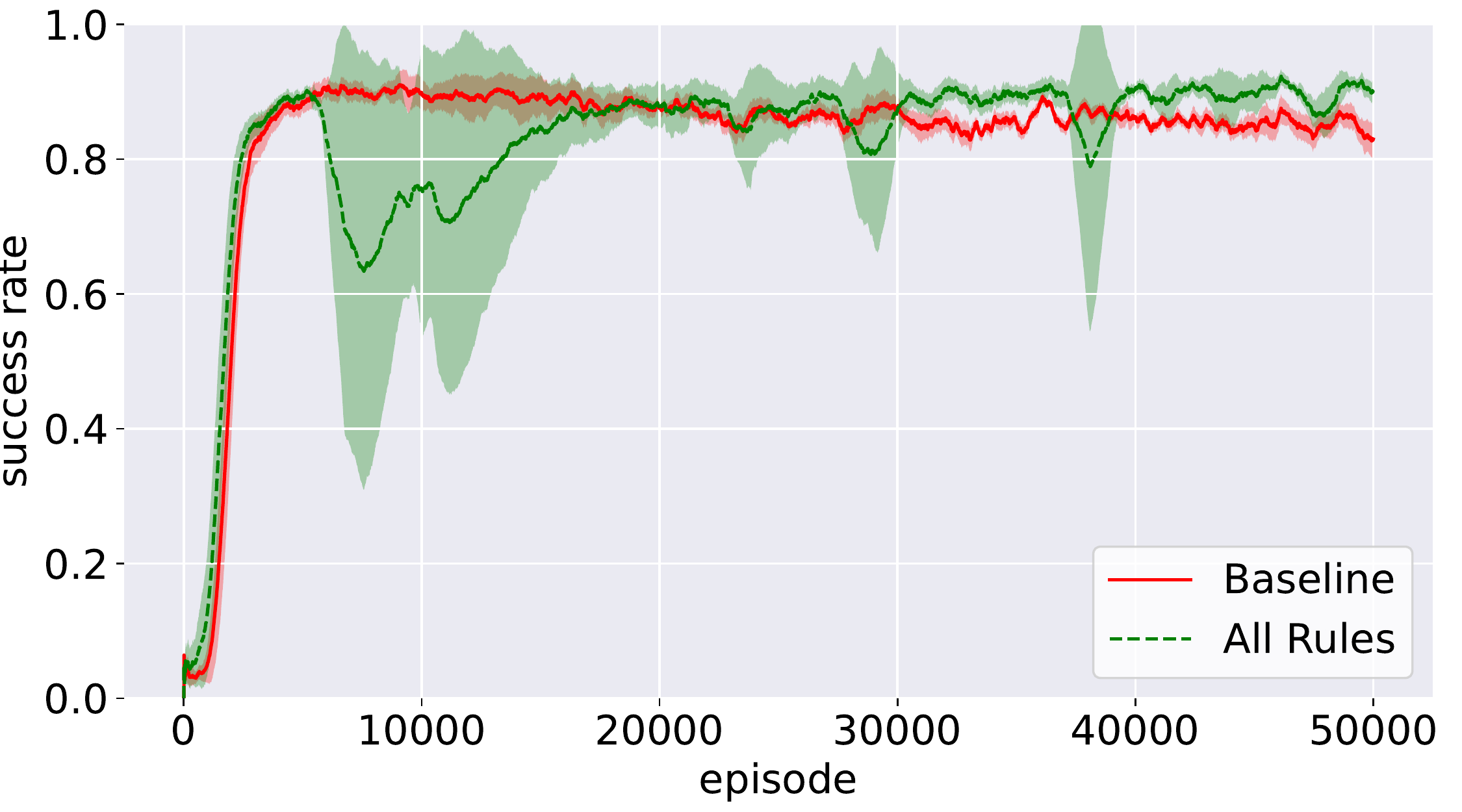}}
    \subfigure[avoid turns larger than 180$^{\circ}$]{\includegraphics[width=0.33\textwidth]{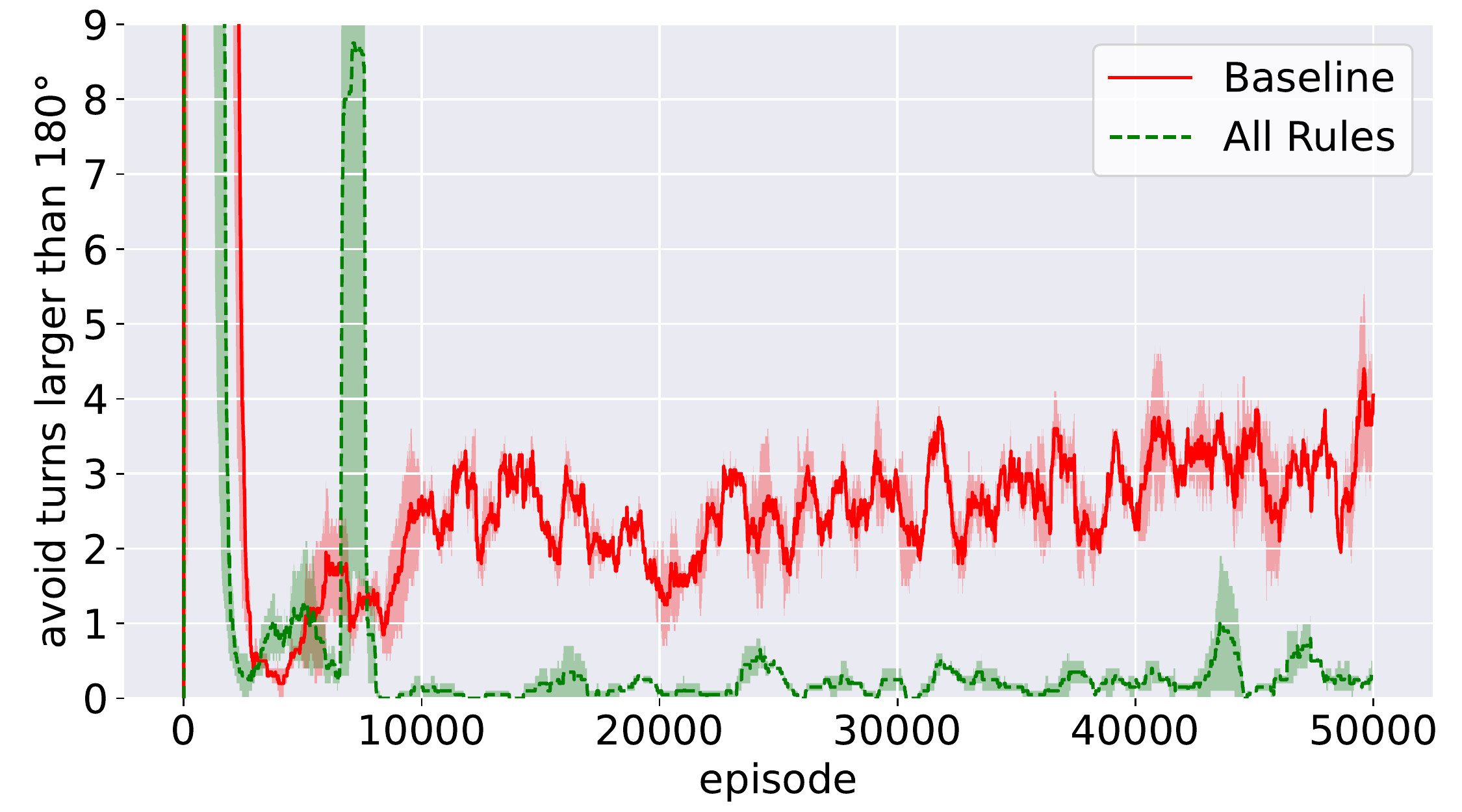}}
    }
  \scalebox{1.0} {
    \subfigure[avoid back-and-forth rotation]{\includegraphics[width=0.33\textwidth]{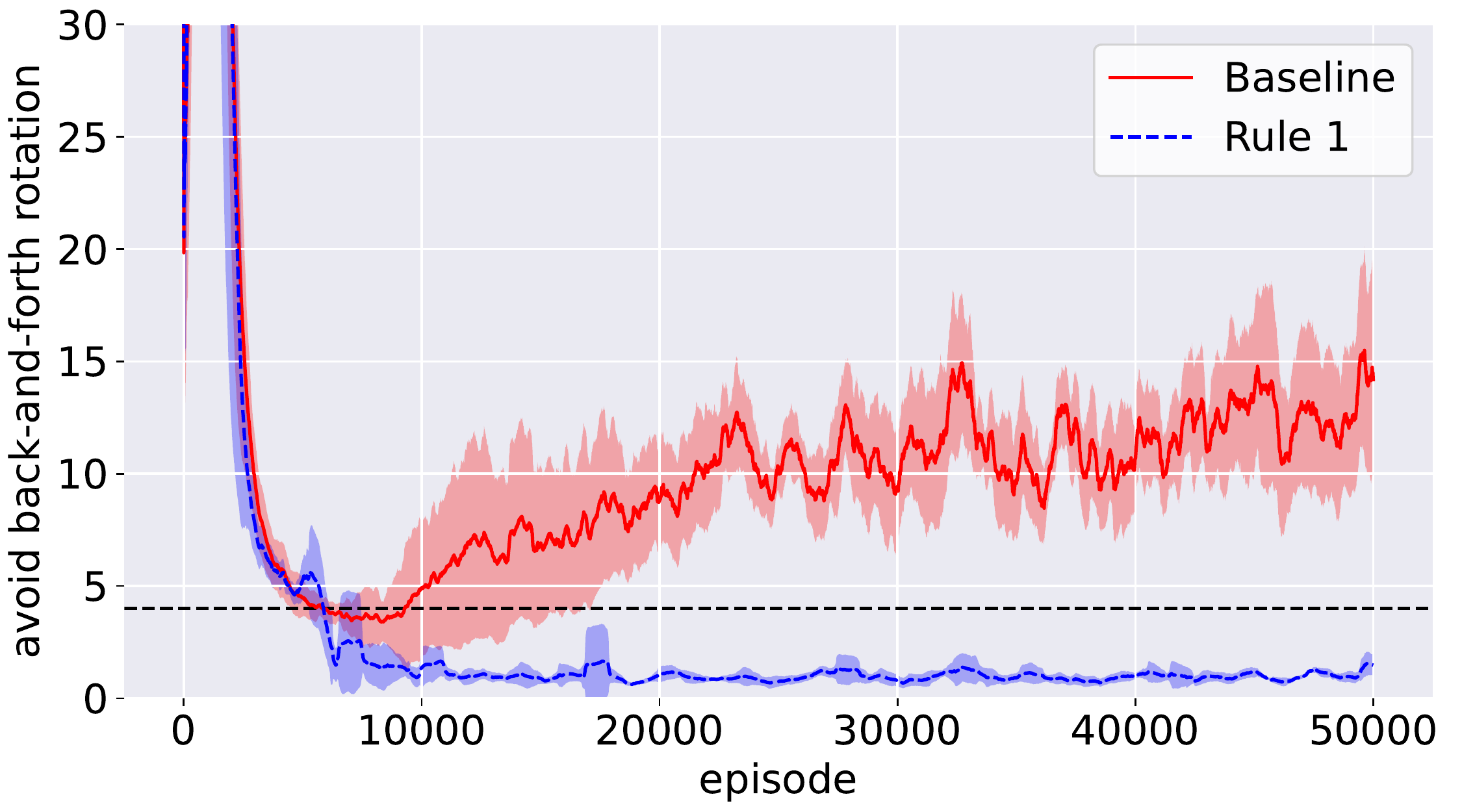}}
    \subfigure[avoid back-and-forth rotation]{\includegraphics[width=0.32\textwidth]{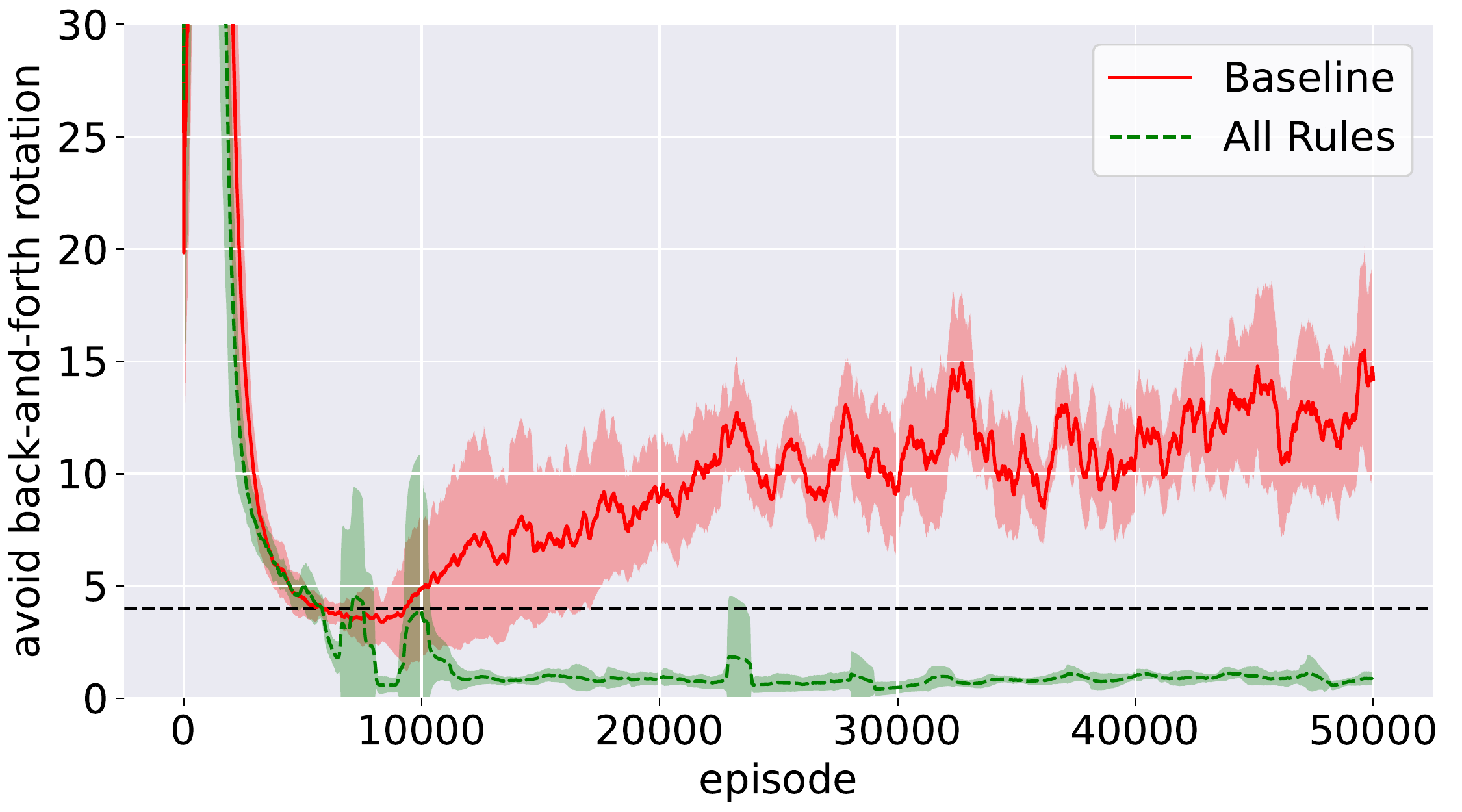}}
    \subfigure[avoid turning when clear]{\includegraphics[width=0.33\textwidth]{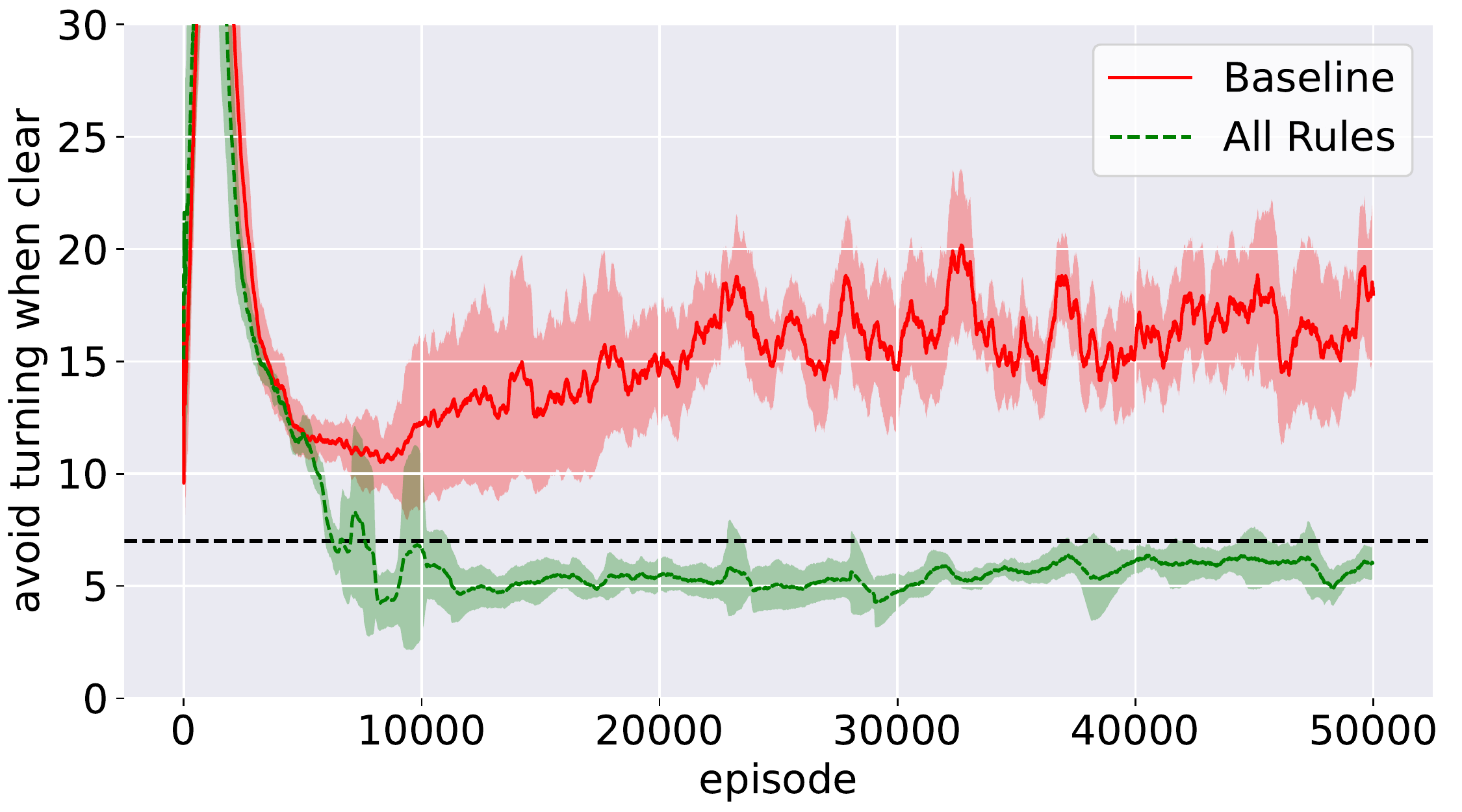}}
  }
  \caption{A comparison between the baseline policies to policies trained using
    our approach. The black dotted line states the threshold ($d_k$) we considered for the $k^{th}$ rule. }
	\label{fig:results:rules}
\end{figure*}

\mysubsection{Results.}
Fig.~\ref{fig:results:rules} depicts a comparison between policies
trained with a standard end-to-end PPO~\citep{ShWoDh17} (the
baseline), and those trained using our constrained method with the
injection of rules.  
In Figs.~\ref{fig:results:rules}(a) and~\ref{fig:results:rules}(d), we show results of  policies trained with just
\scenarioOne~added as a constraint.

Fig.~\ref{fig:results:rules}(a) shows that the
success rate of the baseline stabilizes at around $87\%$, while the success
rate of our improved policies  stabilizes at around
$95\%$. 

Fig.~\ref{fig:results:rules}(d) then compares the frequency of
undesired behavior occurrences between the baseline, at about $13$ per
episode, and our policies, where the frequency 
diminishes \emph{almost completely}.

Next, for Fig.~\ref{fig:results:rules}(b) we show results of  policies trained
with all three of our added rules; we note that the success rate
for these policies stabilizes around $95\%$, compared to $87\%$ for
the baseline.

Finally, in Figs.~\ref{fig:results:rules}(c), (e) and (f), we compare
the frequency of the occurrence of undesired behaviors between the
baseline and the policies trained with all rules active. Using the
baseline, the frequency of the three behaviors is about $13$, $3$, and
$17$ per episode. 
The undesired behaviors are removed \emph{almost completely} for the policies trained with our additional rules and method.

We note that the undesired behavior addressed by the rule \scenarioTwo{}
is quite rare in general; and so the statistics reported in
Fig.~\ref{fig:results:rules}(c) were collected over the final $100$
episodes of training.

The results clearly show that our method is able to train agents that
respect the given constraints, without damaging the main training objective --- the success rate.
Moreover, it also highlights the
scalability of our method, i.e., performing well when single or multiple rules are applied.
Reviewing Fig~\ref{fig:results:rules}(b), comparing the baseline's success rate with our method's success rate, when all rules are applied together with all the optimizations presented in Section~\ref{sec:methods:lagrangian}, shows a clear advantage.

Excitingly, our approach even led to an improved success rate,
suggesting that the contribution of expert knowledge can drive the
training to better policies. This showcases the importance of
enabling expert-knowledge contributions, compared to end-to-end
approaches.

\mysubsection{Formal Verification and Safety Guarantees.}  To further
prove the effectiveness of our method, we show results of using the
Marabou DNN verification engine~\citep{KaHuIbHuLaLiShThWuZeDiKoBa19,
  OsBaKa22, WuZeKaBa22, StWuZeJuKaBaKo21, WuOzZeIrJuGoFoKaPaBa20, AmWuBaKa21}
to assess the reliability of our trained models. DNN verification is a
sound and complete method for checking whether a DNN model displays
unwanted behavior, over \emph{all} possible inputs.

In order to conduct a fair comparison, we
selected only models that passed our success cutoff value (85\%); and
for each of these models we ran three verification queries --- each
checking whether the model violates a given property (\sat), or abides
by it for all inputs (\unsat). We note that a verifier might also fail
to terminate, due to \timeout{} or \memout{} errors.  Each query ran
with a \timeout{} value of $36$ hours, and a \memout{} value of $6$ GB.
Table~\ref{tab:results:verification_appendix} summarizes the results
of our experiments.

These results show a \emph{significant} change of behavior between
DNNs trained with the baseline algorithm, and those trained by our
method. Indeed, we see that the latter policies much more often
completely abide by the specific rules, and are consequently far more
reliable.

\begin{table}[ht]
\small
\caption{Results of the formal verification queries over a total of 120
  trained DNNs, for each of the three properties in question.  The
  first row shows the results of the 60 baseline policies, and the second
  row shows results of the 60 policies trained by our method, with all
  rules active.}
\vspace{0.3cm}
\begin{tabular}{l|ccc|ccc|ccc|}
\multicolumn{1}{c|}{} & \multicolumn{3}{c|}{\scenarioOne} & \multicolumn{3}{c|}{\scenarioTwo} & \multicolumn{3}{c|}{\scenarioThree} \\
\midrule
ALGO & \sat & \unsat & \timeout & \sat & \unsat & \timeout & \sat & \unsat & \timeout \\
\midrule
Baseline & 60 & 0 & 0 & 51 & 0 & 9 & 60 & 0 & 0 \\
SBP & 22 & 38 & 0 & 0 & 41 & 19 & 9 & 34 & 17 \\
\end{tabular}
\vspace{5pt}
\label{tab:results:verification_appendix}
\end{table}
\section{Related Work}
\label{sec:related_work}

To the best of our knowledge, this is the first work that combines
scenario-based programming into the training of a constrained
reinforcement learning system --- specifically in a robotic environment.

In~\citep{Yerushalmi2022ScenarioAssistedDR}, the authors proposed an
integration between SBP and DRL training, using a reward shaping
approach that penalizes the agent when rules are violated. 
\begin{figure*}[ht]
	\centering
	\subfigure[success rate]{\includegraphics[width=0.32\textwidth]{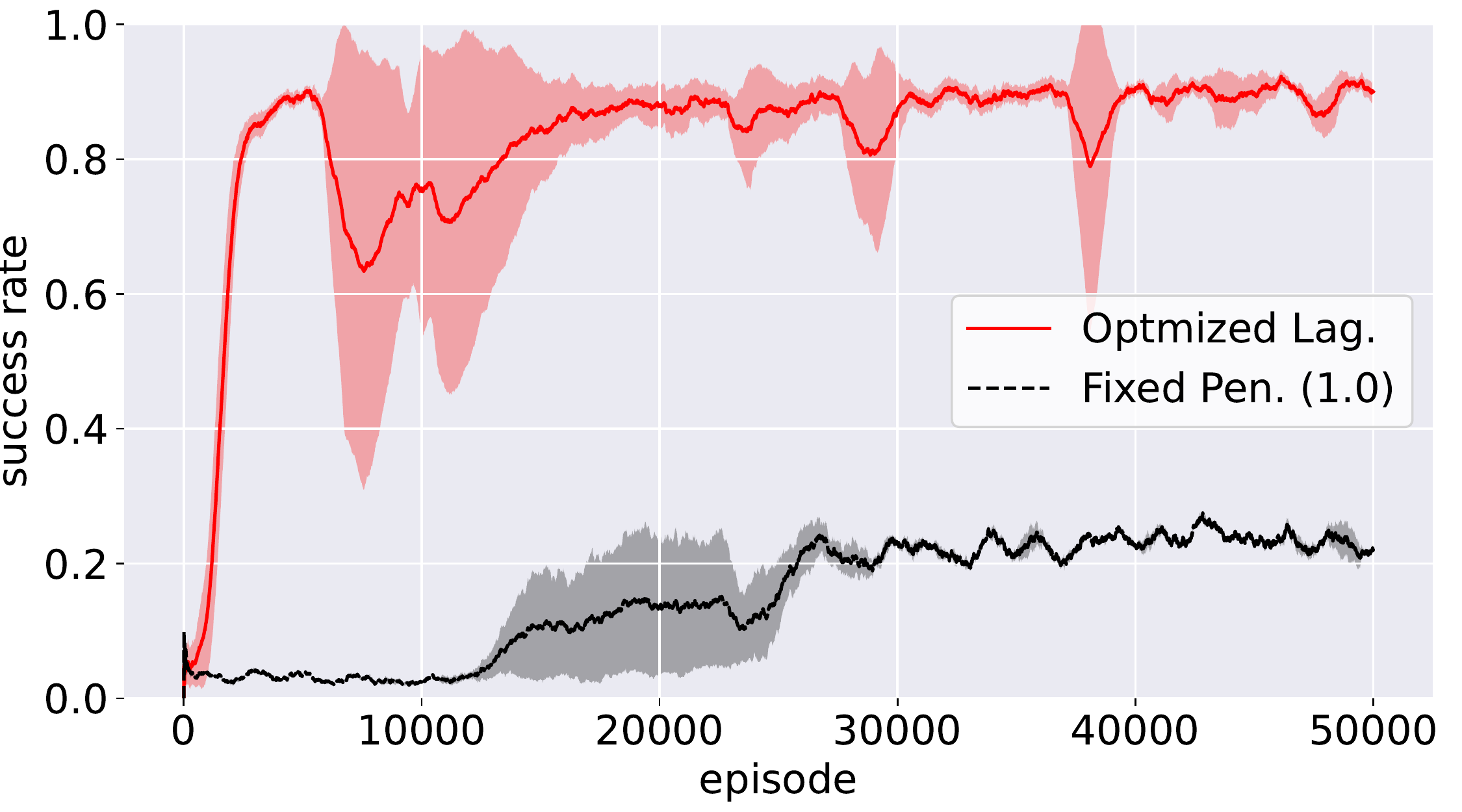}}
	\subfigure[avoid back-and-forth rotation]{\includegraphics[width=0.33\textwidth]{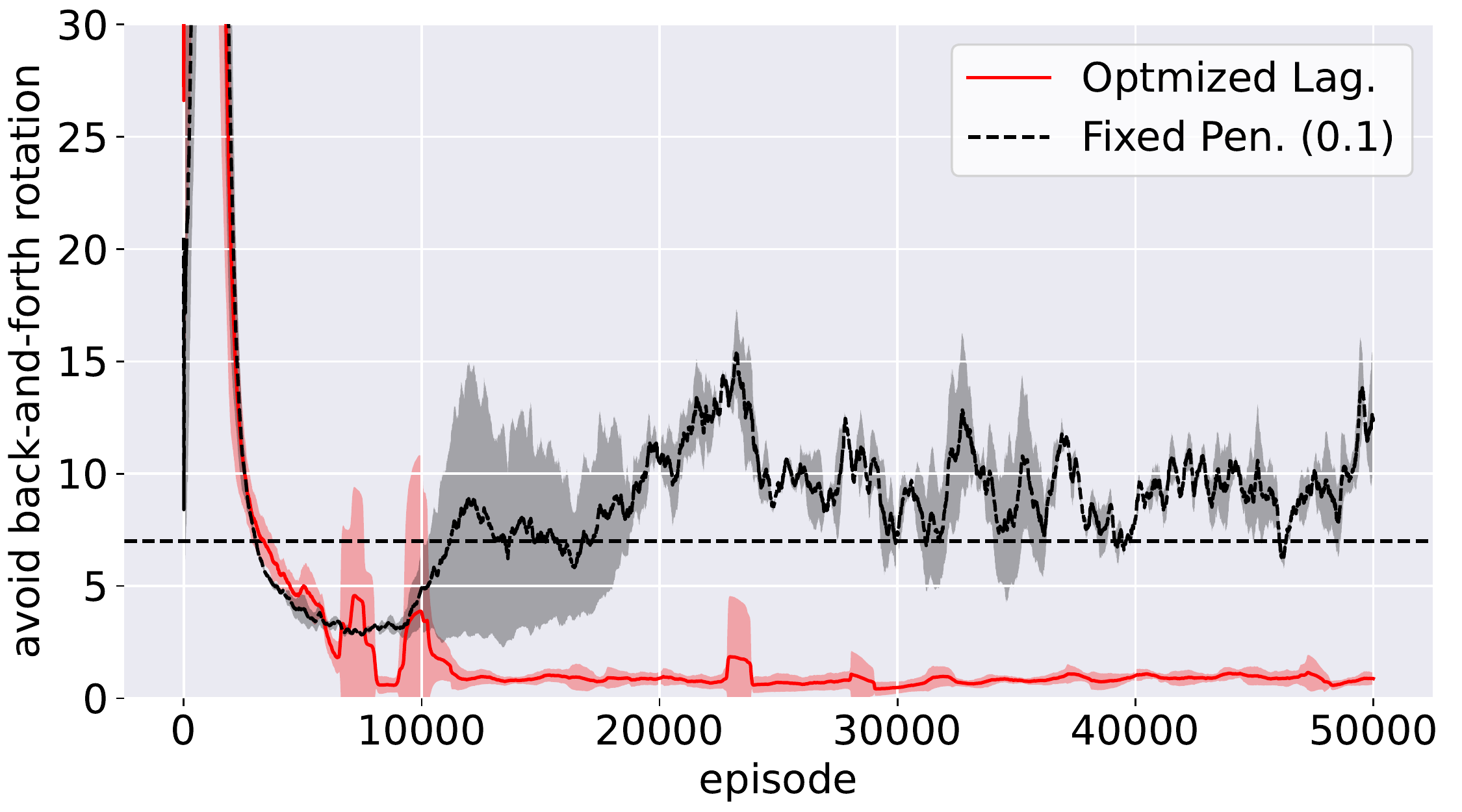}}
	\subfigure[success rate]{\includegraphics[width=0.32\textwidth]{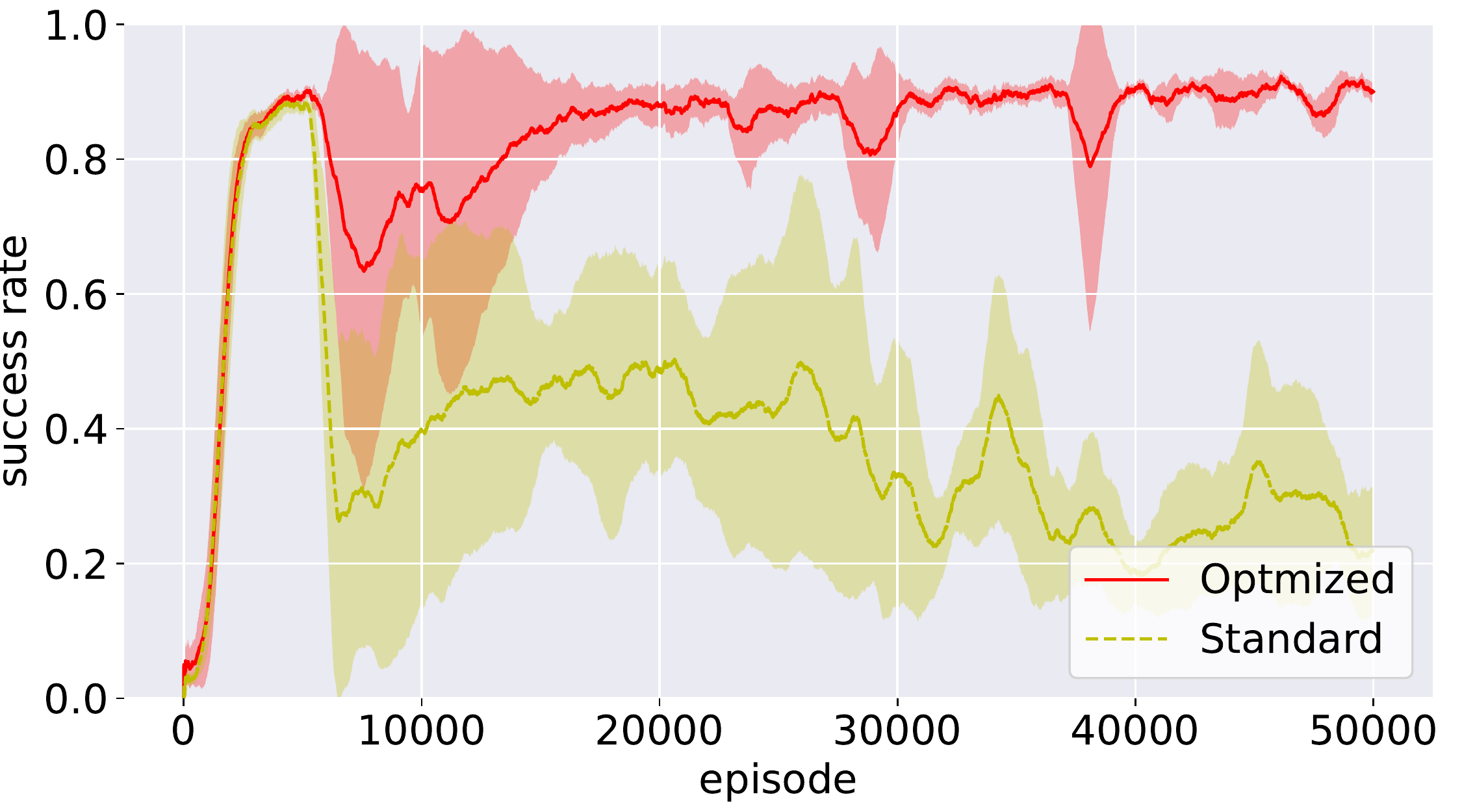}}
	
	\caption{Graphs (a) and (b) show a comparison of our approach
          with that of~\citep{Yerushalmi2022ScenarioAssistedDR}: 
          graph (a) compares the success rates of the two approaches
          with all three scenario-based rules, and graph (b) compares the
          frequency of violations to the \scenarioOne{} rule. For
          these two experiments, we configured the approach
          of~\citep{Yerushalmi2022ScenarioAssistedDR} to use fixed
          penalties of $1.0$ and $0.05$, respectively. 
          Graph (c) compares to the success rate of our approach to
          that of a policy trained with standard Lagrangian-PPO. }
	\label{sec:appendix:related_work:comparison}
\end{figure*} 

Our approach provides agents with fewer rule violations; parts (a) and (b) of Fig.~\ref{sec:appendix:related_work:comparison}
depict a comparison between our approach and that of~\citet{Yerushalmi2022ScenarioAssistedDR}, using different reward
penalties to compare their effectiveness.
Although the two approaches share some traits, their work requires us to manually
determine the penalty that is incurred whenever the agent violates the
scenario-based rules --- which can be quite
difficult~\citep{Yerushalmi2022ScenarioAssistedDR}. Furthermore, this
limitation renders the approach more difficult to apply incrementally:
each additional scenario that is added to the SB program might require
re-adjustments of the reward penalties, and this might become highly
difficult for a large number of scenarios.

Constrained reinforcement learning is an emerging
field~\citep{LiDiLi20, MaCoFa21b, AcHeTa17}. To show the effectiveness
of our approach, we also compared it to an implementation of
Lagrangian-PPO, as suggested by~\citep{RaAcAm19}. The comparison
results are shown in Fig.~\ref{sec:appendix:related_work:comparison}
(c).  Although the technique of~\citep{RaAcAm19} is able to reduce the
number of violations, it fails to reach a high success rate.

In a recent work on constrained reinforcement
learning~\citep{RoGiRo21}, the authors advocate an optimized version
of Lagrangian-PPO. They propose a different approach to balance the
constraints and the return, based on the softmax activation function
and without imposing bounds on the values for the
multipliers. Moreover, their work focuses on a different domain (game
development), which presents very different challenges compared to
robotics (e.g., safety and efficiency are not considered as crucial
requirements); and they do not encode constraints using a framework
geared for this purpose, such as SBP.

\mysubsection{Limitations.} Our method suffers from various
limitations. First, it does not completely guarantee that the
resulting policies are safe. For example, as shown in Table ~\ref{tab:results:verification_appendix}. 
Even though the number of
formally safe models is significant, it is not absolute. In addition, using verification to check this may not always be
feasible due to various limitations of current verification
technology. 

Second, our method requires prior knowledge of 
scenario-based programming to formalize the properties. To
mitigate this, our approach can be extended to support additional
rule-specifying formalisms, in addition to SBP. Third, the scalability
of the method needs to be investigated. We showed in this work that
the algorithm can easily handle one to three scenarios, in addition to
the main objective. We leave to future work the analysis of
performance when the number of properties increases further.
\section{Conclusion}
\label{sec:conclusion}

This paper presents a novel and generic approach for incorporating
subject-matter-expert knowledge directly into the DRL learning
process, allowing to achieve user-defined safety properties and
behavioral requirements.  We show how to encode the desired behavior
as constraints for the DRL algorithm and improve a state-of-the-art
algorithm with various optimizations.  Importantly, we define
properties comprehensibly, leveraging scenario-based
programming to encode them into the training loop.  We apply our
method to a real-world robotic problem, namely mapless navigation, and
show that our method can produce policies that respect all the
constraints without adversely affecting the main objective of the
optimization. We further demonstrate the effectiveness of our method
by providing formal guarantees, using DNN verification, about the
safety of trained policies.

Moving forward, we plan to extend our work to different environments including navigation in more complex domains (e.g., air and
water). Another key challenge for the future is to inject rules aiming
to encode behaviours in a cooperative (or competitive) multi-agent
environment.

\section*{Acknowledgements}
The work of Yerushalmi, Amir and Katz was partially supported by the
Israel Science Foundation (grant numbers 683/18 and 3420/21) and the
Israeli Smart Transportation Research Center (ISTRC). The work of
Corsi and Farinelli was partially supported by the ``Dipartimenti di
Eccellenza 2018--2022'' project, and funded by the Italian Ministry of
Education, Universities and Research (MIUR). The work of Harel and
Yerushalmi was partially supported by a research grant from the Estate
of Harry Levine, the Estate of Avraham Rothstein, Brenda Gruss and
Daniel Hirsch, the One8 Foundation, Rina Mayer, Maurice Levy, and the
Estate of Bernice Bernath.

\bibliography{ScenarioBasedDRL_CoRL2022}

\clearpage
\appendix
\section{SBP Python Objects Implementation}
\label{sec:appendix:sbm}

The Python implementation of the three scenarios used in this paper is
shown below: the code for \scenarioOne{} appears in
Fig.~\ref{fig:code:scenario-1}, the code for \scenarioTwo{} appears in
Fig.~\ref{fig:code:scenario-2}, and the code for \scenarioThree{}
appears in Fig.~\ref{fig:code:scenario-3}.

\begin{figure}[ht]
\begin{lstlisting}[language=Python]
def SBP_avoidBackAndForthRotation():
    blockedEvList = []
    waitforEvList = [BEvent("SBP_MoveForward"),
                     BEvent("SBP_TurnLeft"),
                     BEvent("SBP_TurnRight")]
    while True:
        lastEv = yield {waitFor: waitforEvList, block: blockedEvList}
        if lastEv != BEvent("SBP_TurnLeft")  
            and lastEv != BEvent("SBP_TurnRight"):
            blockedEvList = []
        else:
            blocked_ev = BEvent("SBP_TurnRight") 
                if lastEv == BEvent("SBP_TurnLeft")
                else BEvent("SBP_TurnLeft")
            # Blocking!
            blockedEvList.append(blocked_ev)
\end{lstlisting}
  \caption{The Python implementation of scenario~\scenarioOne. The
    code waits for any of the possible events:
    \emph{SBP\_MoveForward, SBP\_TurnLeft} and
    \emph{SBP\_TurnRight}. Upon receiving \emph{SBP\_TurnLeft}, it
    blocks \emph{SBP\_TurnRight}, and upon receiving
    \emph{SBP\_TurnRight}, it blocks \emph{SBP\_TurnLeft}.  Upon
    receiving \emph{SBP\_MoveForward}, it clears any blocking.}
\label{fig:code:scenario-1}
\end{figure}

\begin{figure}[ht]
\begin{lstlisting}[language=Python]
def SBP_avoid_k_consecuative_turns():
    k = 7
    counter = 0
    prevEv = None
    blockedEvList = []
    waitforEvList = [BEvent("SBP_MoveForward"), BEvent("SBP_TurnLeft"), \\
    BEvent("SBP_TurnRight")]
    while True:
        lastEv = yield {waitFor: waitforEvList, block: blockedEvList}
        if prevEv is None or lastEv == BEvent("SBP_MoveForward") or prevEv != lastEv:
            prevEv = lastEv
            counter = 0
            blockedEvList = []
        else:
            if counter == k - 1:
                # Blocking!
                blockedEvList.append(lastEv)
            else:
                counter += 1
\end{lstlisting}
  \caption{The Python implementation of a scenario that blocks turning
    in the same direction more then k consecutive times. Each turn
    action rotates the robot by $30^{\circ}$, and so we set k to be 7. }
\label{fig:code:scenario-2}
\end{figure}

\begin{figure}[t]
\begin{lstlisting}[language=Python]
def SBP_avoid_turning_when_clear():
    blockedEvList = []
    waitforEvList = [BEvent("SBP_MoveForward"), BEvent("SBP_TurnLeft"),\\
    BEvent("SBP_TurnRight")]
    while True:
        lastEv = yield {waitFor: waitforEvList, block: blockedEvList}
        state = lastEv.data['state']
        if state[3] > MINIMAL_FWD_CLEARANCE and state[2] > MINIMAL_CLEARANCE and \\
                state[4] > MINIMAL_CLEARANCE and abs(FWD_DIR - state[-2]) < FWD_DIR_TOLERANCE: 
            blockedEvList.extend([BEvent("SBP_TurnLeft"), BEvent("SBP_TurnRight")])
        else:
            blockedEvList = []
\end{lstlisting}
  \caption{The Python implementation of a scenario that blocks turning
    if the target is straight ahead and the path towards it is
    clear. The event carries data with it, which includes readings
    from the seven lidar sensors --- with state[3] being the front-heading
    sensor. State[-2] is the direction to the target.}
\vspace*{10in}
\label{fig:code:scenario-3}
\end{figure}


\end{document}